# Effect of Different Distance Measures on the Performance of K-Means Algorithm: An Experimental Study in Matlab

Dibya Jyoti Bora, Dr. Anil Kumar Gupta

*Department Of Computer Science & Applications,*

*Barkatullah University, Bhopal, India*

**Abstract:** K-means algorithm is a very popular clustering algorithm which is famous for its simplicity. Distance measure plays a very important rule on the performance of this algorithm. We have different distance measure techniques available. But choosing a proper technique for distance calculation is totally dependent on the type of the data that we are going to cluster. In this paper an experimental study is done in Matlab to cluster the iris and wine data sets with different distance measures and thereby observing the variation of the performances shown.

## 1. INTRODUCTION:

Clustering is an unsupervised study. The main aim of clustering is to divide a dataset into some different subsets (known as 'clusters') such that data into a particular subset sharing similar properties while data in different subset showing different properties from data in another subset. Means to say that clustering should satisfy the two properties [1][2]: 1. High Cohesive Property and 2. Low Coupling Property. Clustering algorithms are mainly divided into two types based on developed cluster properties: hierarchical and partitional. The hierarchical methods, in general try to decompose the dataset of n objects into a hierarchy of groups [1]. This hierarchical decomposition can be represented by a tree structure diagram called as a dendrogram [3]; whose root node represents the whole dataset and each leaf node is a single object of the dataset. The clustering results can be obtained by cutting the dendrogram at different level. There are two general approaches for the hierarchical method: agglomerative and divisive [2][3][4]. Agglomerative approach is a bottom up approach starting with n-leaf nodes, letting them as individual clusters, moving upwards towards the root with some merging criteria. While divisive hierarchical clustering technique is a top down approach, starting from the root node gradually splitting the data into different clusters downwards based on the properties of the data.

While in case of partitional clustering, k partitions of the datasets with n objects are created, where each partition represents a cluster, where k<= n. It tries to divide the data into subsets or partitions based on some evaluation criteria. K-Means is a very popular partitional clustering algorithm. In this paper, first of all the K-Means algorithm is discussed and then different distance measurement techniques for K-means algorithm are mentioned. After that we go through experiments for implementing K-Means with different distance measurement techniques in Matlab. Based on the results of the experiments, we finally come to a conclusion about the performance of K-Means algorithm with respect to different distance measurement techniques.

## 2. K –MEANS ALGORITHM:

The K-Means [5] is one of the famous partition clustering algorithm [ 6][7][8]. It takes the input parameter *k*, the number of clusters, and partitions a set of *n* objects into *k* clusters so that the resulting intra-cluster similarity is high but the inter-cluster similarity is low. The main idea is to define k centroids, one for each cluster. These centroids should be placed in a cunning way because of different location causes different results. So, the better choice is to place them as much as possible far away from each other. The next step is to take each point belonging to a given data set and associate it to the nearest centroid. When no point is pending, the first step is completed and an early groupage is done. At this point we need to re-calculate k new centroids. After we have these k new centroids, a new binding has to be done between the same data set points and the nearest new centroid. A loop has been generated. As a result of this loop we may notice that the k centroids change their location step by step until no more changes are done. In other words centroids do not move any more. Finally, this algorithm aims at minimizing an *objective function*, in this case a squared error function. The objective function

$$J = \sum_{j=1}^{k}\sum_{i=1}^{n}\left\| x_i^{(j)} - c_j \right\|^2$$

The Formal Algorithm [8] is :

1. Select K points as initial centroids.
2. Repeat.
3. Form k clusters by assigning all points to the closest centroid.
4. Recompute the centroid of each cluster. Until the centroids do not change

A diagrammatic view[9] of the K-Means algorithm is :

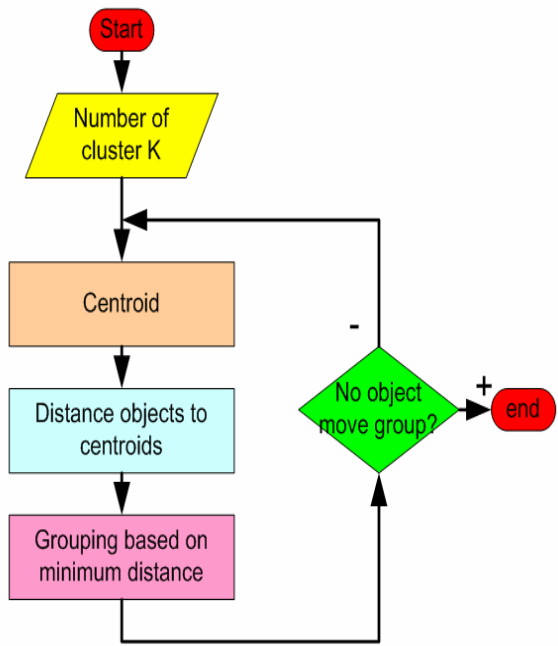

*Figure (1): K-Means Algorithm*

### 3. DISTANCE MEASUREMENTS IN K-MEANS ALGORITHMS:

In K-Means algorithm, we calculate the distance between each point of the dataset to every centroid initialized. Based on the values found, points are assigned to the centroid with minimum distance. Hence, this distance calculation plays the vital role in the clustering algorithm. As we know, distance between two points can be computed with different techniques available, so, our main aim is to pick up a proper technique from the available ones. But, in choosing such techniques, some important points to be noted such as: the property of the data and the dimension of the dataset. In this experiment, we take "Cityblock","Euclidean","Cosine" and "Correlation"-these distance measurement techniques for distance calculations in the K-Means algorithm. Description about each technique is mentioned below:

### 3.1 City Block (Manhattan):
The city block distance [10][11] two point a and b with k dimensions is defined as:

$$\sum_{j=1}^{k} |a_j - b_j|$$

The name *City block distance* (also referred to as *Manhattan distance*) [11] is explained if we consider two points in the xy-plane. The shortest distance between the two points is along the hypotenuse, which is the *Euclidean distance*. The *City block distance* is instead calculated as the distance in x plus the distance in y, which is similar to the way we move in a city (like Manhattan) where we have to move around the buildings instead of going straight through.

### 3.2 Euclidean distance:
The Euclidean distance between two points, *a* and *b*, with *k* dimensions is calculated as[12][13]:

$$\sqrt{\sum_{j=1}^{k}(a_j - b_j)^2}$$

Euclidean (and squared Euclidean) distances are usually computed from raw data, and not from standardized data. One advantage of this method is that the distance between any two objects is not affected by the addition of new objects to the analysis, which may be outliers [13]. However, the distances can be greatly affected by differences in scale among the dimensions from which the distances are computed. For example, if one of the dimensions denotes a measured length in centimeters, and you then convert it to millimeters (by multiplying the values by 10), the resulting Euclidean can be greatly affected (i.e., biased by those dimensions which have a larger scale), and consequently, the results of cluster analyses may be very different. Generally, it is good practice to transform the dimensions so that they have similar scales [13].

### 3.3 Cosine Distance:
The cosine distance between two points is one minus the cosine of the included angle between points (treated as vectors). Given an *m*-by-*n* data matrix X, which is treated as *m* (1-by-*n*) row vectors $x_1, x_2, ..., x_m$, the cosine distances between the vector $x_s$ and $x_t$ are defined as follows[14]:

$$d_{st} = 1 - \frac{x_s x_t'}{\sqrt{(x_s x_s')(x_t x_t')}}$$

3.4 Correlation Distance:

Distance correlation is a measure of dependence between random vectors [15]. Given an *m*-by-*n* data matrix X, which is treated as *m* (1-by-*n*) row vectors $x_1, x_2, ..., x_m$, the correlation distances between the vector $x_s$ and $x_t$ are defined as follows[14]:

$$d_{st} = 1 - \frac{(x_s - \bar{x}_s)(x_t - \bar{x}_t)'}{\sqrt{(x_s - \bar{x}_s)(x_s - \bar{x}_s)'}\sqrt{(x_t - \bar{x}_t)(x_t - \bar{x}_t)'}}$$

### 4. EXPERIMENTS:
For our experiments, we have chosen two different datasets: iris and wine datasets. Iris dataset The Iris flower data set or Fisher's Iris data set (some times also known as Anderson's Iris data) is a multivariate data set introduced by Sir Ronald Fisher (1936) as an example of discriminant analysis. The data set consists of 50 samples from each of three species of Iris (Iris setosa, Iris virginica and Iris versicolor). Four features were measured from each sample: the length and the width of the sepals and petals, in centimeters[16]. Table(1) gives details of the iris dataset[16]:

| Data Set Characteristics: | Multivariate | Number of Instances: | 150 | Area: | Life |
|---|---|---|---|---|---|
| Attribute Characteristics: | Real | Number of Attributes: | 4 | Date Donated | 1988-07-01 |
| Associated Tasks: | Classification | Missing Values? | No | Number of Web Hits: | 548538 |

*Table (1): Iris dataset*

| Data Set Characteristics: | Multivariate | Number of Instances: | 178 | Area: | Physical |
|---|---|---|---|---|---|
| Attribute Characteristics: | Integer, Real | Number of Attributes: | 13 | Date Donated | 1991-07-01 |
| Associated Tasks: | Classification | Missing Values? | No | Number of Web Hits: | 328827 |

*Table (2) Wine dataset*

The wine dataset [17] is the result of a chemical analysis of wines grown in the same region in Italy but derived from three different cultivars. The analysis determined the quantities of 13 constituents found in each of the three types of wines. Table (2) gives details of the wine dataset [17]:

We have "kmeans" function to perform K-means clustering in Matlab [18]. The function kmeans performs K-Means clustering, using an iterative algorithm that assigns objects to clusters so that the sum of distances from each object to its cluster centroid, over all clusters, is a minimum. k means returns an n-by-1 vector IDX containing the cluster indices of each point. When X is a vector, kmeans treats it as an *n*-by-1 data matrix, regardless of its orientation.

4.1 Experiments with Iris dataset:

We specify the K (number of clusters) as 3. Results for running K-Means algorithm for iris dataset with different distance measures are mentioned below:

(A) City Block distance:
```
 iter    phase    num     sum
  1       1      150     1679
  2       1       5      1651
  3       1       1      1649
  4       2       0      1649
```
Best total sum of distances = 1649

(B) Euclidean distance:
```
iter     phase    num      sum
 1        1      150     14762.1
 2        1       10     14360.7
 3        1       31      8975.18
 4        1       13      7989.84
 5        1        3      7941.63
 6        1        2      7899.12
 7        2        0      7897.88
```
Best total sum of distances = 7897.88

(C) Cosine Distance:
```
iter     phase    num      sum
 1        1      150     0.161957
 2        1        1     0.161835
 3        2        0     0.161835
```
Best total sum of distances = 0.161835

(D) Correlation Distance:
```
 iter    phase      num        sum
  1       1        150      4.61083
  2       1         50      0.463321
  3       1          4      0.371327
  4       1          1      0.370903
  5       2          0      0.370903
```
Best total sum of distances = 0.370903

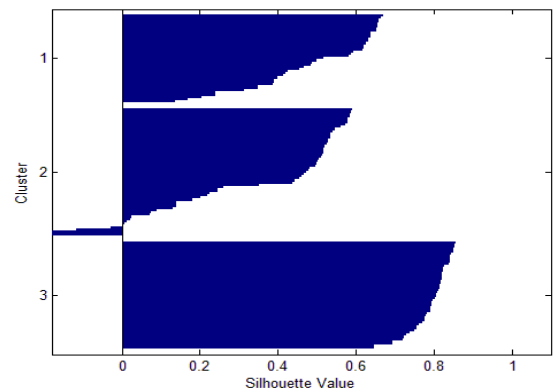

*Figure (2.1) Cluster figure for City block distance*

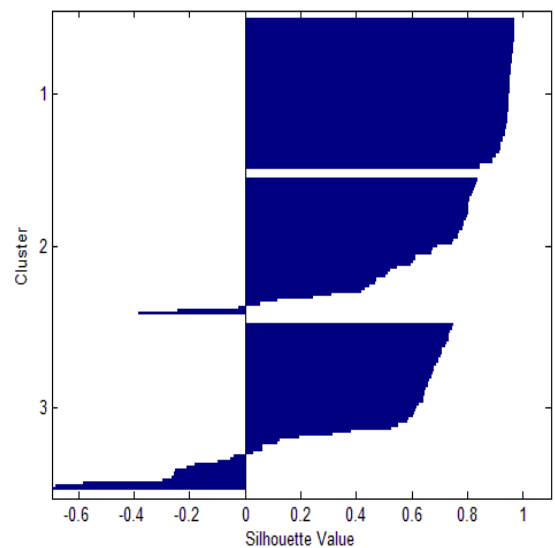

*Figure (2.2) Cluster figure for Euclidean distance*

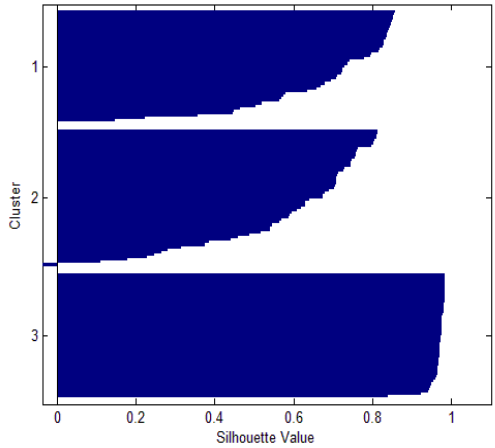

*Figure (2.3) Cluster figure for Cosine distance*

(C) Cosine Distance:

| iter | phase | num | sum |
|---|---|---|---|
| 1 | 1 | 13 | 0.321896 |
| 2 | 1 | 1 | 0.296767 |
| 3 | 1 | 1 | 0.281313 |
| 4 | 2 | 1 | 0.277909 |
| 5 | 2 | 0 | 0.273201 |

Best total sum of distances = 0.273201

(D) Correlation Distance:

| iter | phase | num | sum |
|---|---|---|---|
| 1 | 1 | 13 | 5.19803 |
| 2 | 1 | 1 | 4.8358 |
| 3 | 1 | 1 | 4.23239 |
| 4 | 1 | 1 | 3.59637 |
| 5 | 1 | 1 | 3.25178 |
| 6 | 2 | 0 | 3.19287 |

Best total sum of distances = 3.19287

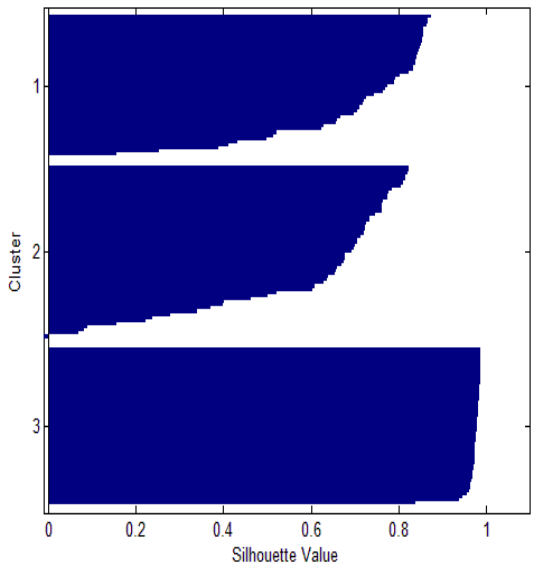

*Figure (2.4) Cluster figure for Correlation distance*

### 4.2 Experiment with Wine dataset:

The results found when running K-Means on Wine datasets with different measures are mentioned below:

(A) City Block Distance:

| iter | phase | num | sum |
|---|---|---|---|
| 1 | 1 | 13 | 155074 |
| 2 | 1 | 4 | 148285 |
| 3 | 1 | 2 | 146085 |
| 4 | 1 | 2 | 121215 |
| 5 | 1 | 1 | 23214.8 |
| 6 | 2 | 0 | 23214.8 |

Best total sum of distances = 23214.8

(B) Euclidean distance:

| iter | phase | num | sum |
|---|---|---|---|
| 1 | 1 | 13 | 860498 |
| 2 | 1 | 2 | 67196 |
| 3 | 2 | 0 | 67196 |

Best total sum of distances = 67196

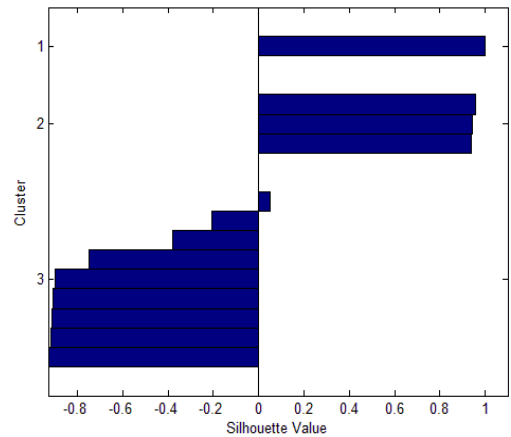

**Figure (3.1) Cluster figure for City block distance**

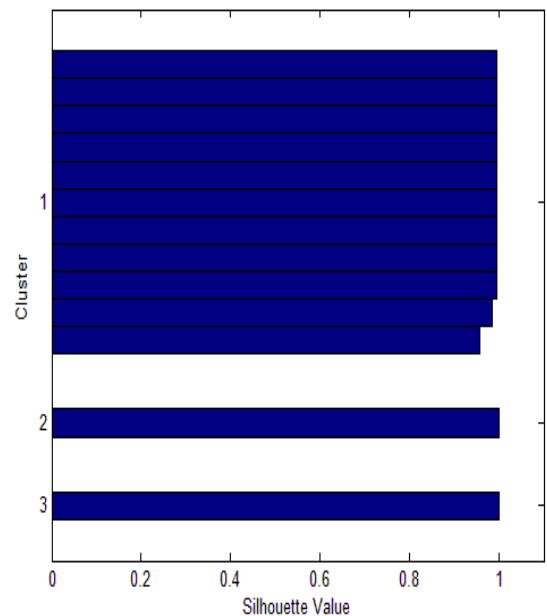

**Figure (3.2) Cluster figure for Euclidean distance**

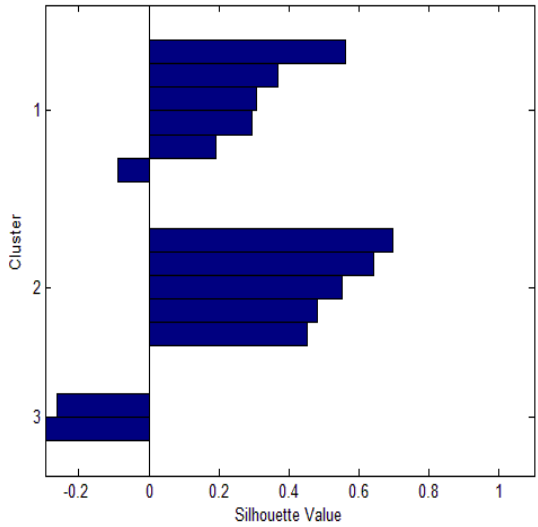

**Figure (3.3) Cluster figure for Cosine distance**

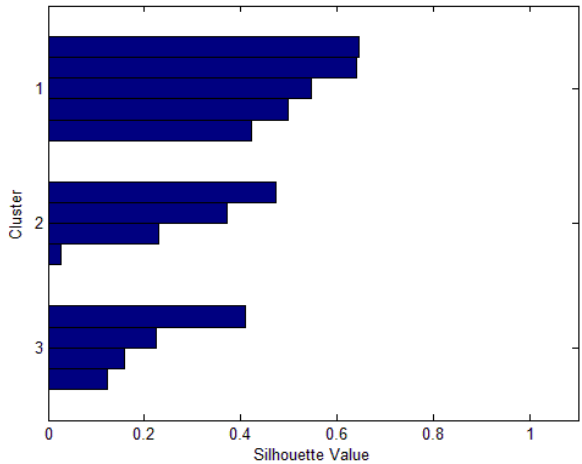

*Figure (3.4) Cluster figure for Correlation distance*

So, from the experiments, we have found different results (in terms of best total sum of distances and clusters' figures) for clustering the same dataset with respect to different distance measures. Also, we have noted down the time taken for each experiment. Table (3) and Table (4) show the respective time taken for the dataset iris and wine respectively:

| City | Euclidean | Cosine | Correlation |
|---|---|---|---|
| 0.069 | 0.079 | 0.091 | 0.085 |

*Table (4): Time with respect to different distance measures for Iris dataset*

| City | Euclidean | Cosine | Correlation |
|---|---|---|---|
| 0.078 | 0.082 | 0.089 | 0.086 |

*Table (5): Time with respect to different distance measures for Wine dataset*

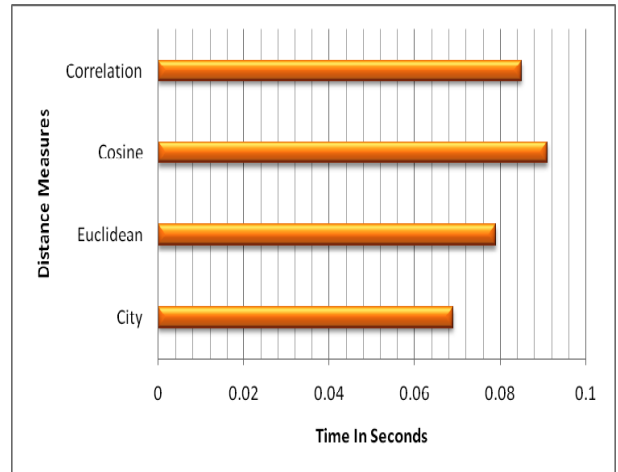

*Figure (4): Comparisons Of different distance measures in terms of computation time for iris dataset*

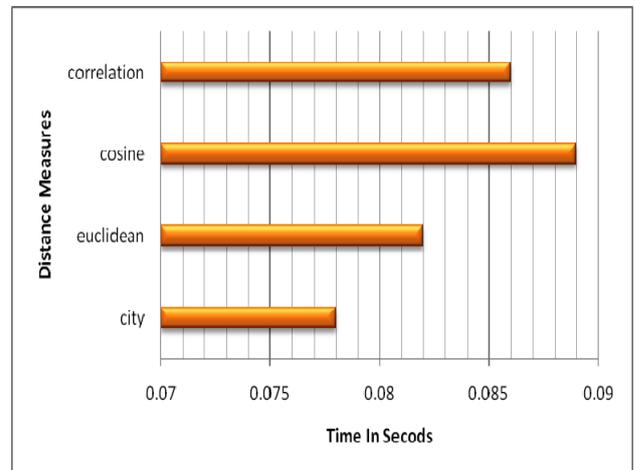

*Figure (5): Comparisons Of different distance measures in terms of computation time for wine dataset*

## CONCLUSION:

On the basis of results of the experiments done, we have found that city block distance shows better performance for both the datasets in terms of less computation time. While cosine takes more computation time in comparison to other distance measures for both the data sets. But, when we refer to silhouette plots of the clusters, then we must admit that "correlation" distance measures show a better interpretation of the clustered data. We have performed our experiments on two different datasets (i.e. iris and wine) only to observe the performance with respect to datasets with different number of attributes. Iris dataset has four attributes, while wine dataset has thirteen attributes. While choosing a proper distance measure, we must also keep an eye on the number of attributes involved in that dataset as otherwise it will result a high time complexity if a wrong one is chosen. In our future work, we will consider another different distance measures for K-Means algorithm with respect to a big dataset and perform a comparison among them, thereby, try to propose a good one for the task of clustering big data set. Also, we will try to extend our study for another partition clustering algorithms like K-Medoids, CLARA and CLARANS.